\documentclass[manuscript,screen]{acmart}

\AtBeginDocument{%
  \providecommand\BibTeX{{%
    \normalfont B\kern-0.5em{\scshape i\kern-0.25em b}\kern-0.8em\TeX}}}

\setcopyright{acmlicensed}
\copyrightyear{2018}
\acmYear{2018}
\acmDOI{XXXXXXX.XXXXXXX}

\usepackage{multirow}
\begin{document}

\title{SiTSE: Sinhala Text Simplification Dataset and Evaluation}

\author{Surangika Ranathunga}
\email{s.ranathunga@massey.ac.nz}
\orcid{0003-0701-0204 }

\affiliation{%
  \institution{School of Mathematical and Computational Sciences, Massey University} 
  \city{Auckland}
  \country{New Zealand}
  \postcode{102904}
}

\author{Rumesh Sirithunga}
\email{rumeshmadhusanka.17@cse.mrt.ac.lk}

\author{Himashi Rathnayake}
\email{himashi.17@cse.mrt.ac.lk}

\author{Lahiru De Silva}
\email{lahirudesilva.17@cse.mrt.ac.lk}

\author{Thamindu Aluthwala}
\email{thamindu.17@cse.mrt.ac.lk}
\affiliation{%
  \institution{Department of Computer Science and Engineering, University of Moratuwa}
  \city{Katubedda}
  \country{Sri Lanka}
  \postcode{10400}}
  
\author{Saman Peramuna}
\email{dmsperamuna@gmail.com}
\affiliation{%
  \institution{Department of Languages, Sabaragamuwa University}
  \country{Sri Lanka}}

\author{Ravi Shekhar}
\email{r.shekhar@essex.ac.uk}
\affiliation{%
  \institution{University of Essex }
  \country{UK}}

\renewcommand{\shortauthors}{Ranathunga, et al.}

\begin{abstract}
  Text Simplification is a task that has been minimally explored for low-resource languages. Consequently, there are only a few manually curated datasets. In this paper, we present a human curated sentence-level text simplification dataset for the Sinhala language. Our evaluation dataset contains 1,000 complex sentences and corresponding 3,000 simplified sentences produced by three different human annotators. We model the text simplification task as a zero-shot and zero resource sequence-to-sequence (seq-seq) task on the multilingual language models mT5 and mBART. We exploit auxiliary data from related seq-seq tasks and explore the possibility of using intermediate task transfer learning (ITTL). Our analysis shows that ITTL outperforms the previously proposed zero-resource methods for text simplification. Our findings also highlight the challenges in evaluating text simplification systems, and support the calls for improved metrics for measuring the quality of automated text simplification systems that would suit low-resource languages as well. Our code and data are publicly available: \url{https://github.com/brainsharks-fyp17/Sinhala-Text-Simplification-Dataset-and-Evaluation}
\end{abstract}

\ccsdesc[500]{Computing methodologies~Artificial intelligence}
\ccsdesc[500]{Computing methodologies~Natural language processing}
\ccsdesc[500]{Computing methodologies~Language resources}
\ccsdesc[500]{Computing methodologies~Language resources}
\keywords{Low-resource languages, sequence-sequence
pre-trained Language Models, mBART, mT5, Transfer Learning,  Intermediate Task Transfer Learning}

\received{xx}
\received[revised]{xx}
\received[accepted]{xx}

\maketitle

\section{Introduction}

Text Simplification (TS) refers to modifying a given text while preserving its original meaning to improve its readability and understanding~\citep{alva-manchego-etal-2020-data}. TS has proven benefits in improving comprehension in low literacy readers~\citep{mason1978facilitating}, helping in learning difficulties~\citep{rello2013frequent}, in teaching L2 learners~\citep{crossley2007linguistic} and for communicating domain-specific information such as medical text to laymen~\citep{garbacea2021explainable}. Thus, TS increases the fairness of textual information processing systems~\citep{garbacea2021explainable}. TS can also be used to improve the performance of other Natural Language Processing (NLP) tasks such as Machine Translation~\citep{agrawal2019controlling} and Summarization~\citep{siddharthan2004syntactic}.


Despite their many benefits, text simplification systems have been implemented mainly for high-resource languages such as English, French and Spanish, except for some work on Urdu, Bengali, Marathi, Basque and Slovene~\citep{anees2021automatic, joseph2023multilingual, mondal2024dimsim}. The main reason for this is due to the lack of complex-simple word lexicons and/or semantic/syntactic parsers used for the early TS systems. Even with the recent focus on low-resource NLP~\citep{schwartz-etal-2019-bootstrapping}, very less attention is paid to the TS task due to the unavailability of datasets and models. Creating high-quality and large TS datasets\footnote{A parallel corpus, where each entry resembles a complex sentence and its simple version} is costly and time-consuming since humans have to read, comprehend, identify possible simplification points and relevant operations, and finally carry out the simplification. Some datasets were automatically curated using Wikipedia to reduce dataset creation costs, but these datasets have not been very useful for TS~\citep{alva-manchego-etal-2020-asset}. Thus, human-curated TS datasets still require language-specific attention.

In this paper, we focus on Sinhala TS. Sinhala is an Indo-Aryan language with its own script and alphabet used in the island nation of Sri Lanka, spoken by about 22M people. Sinhala is a low-resource language due to the scarcity of digital language resources and lack of collaborative research~\citep{de2021survey, ranathunga2022some}. We present the first-ever sentence-level Sinhala Text Simplification Evaluation (SiTSE) dataset. Our dataset consists of 1000 complex sentences from Sinhala official government documents. Each complex sentence has three simple references produced by Sinhala language experts, thus resulting in 3000 sentence pairs. 

The state-of-the-art solutions for TS are based on Deep Learning techniques, in particular sequence-sequence pre-trained Language Models (sqPLMs)~\citep{martin2021muss, joseph2023multilingual}. Semi and unsupervised neural methods have been used to tackle the data scarcity problem. However, the existing semi- and unsupervised neural TS systems~\citep{surya2019unsupervised, zhao2020semi} have mainly been tested using high-resource languages in a simulated low-resource setting. For example~\citet{surya2019unsupervised, zhao2020semi} used the English Wikipedia,~\citet{martin2021muss} used the CommonCrawl of English, French, and Spanish and~\citet{mallinson2020zero} used 6M parallel sentences for English-German.

Sinhala has a fraction of these monolingual and multilingual data. On the positive side, Sinhala has been included in sqPLMs such as mBART~\citep{mbart} and mT5~\citep{xue-etal-2021-mt5}. Despite Sinhala being under-represented in these models, their use has shown very promising results for related sequence-sequence tasks such as Translation and auto-regressive text generation, in the context of Sinhala~\citep{thillainathan2021fine, lee2022pre, niyarepola2022math}. Sinhala also has small amounts of cleaned parallel data~\citep{fernando2020data}, as well as relatively large amounts of web-mined noisy parallel data~\citep{ranathunga2024quality}.

We experiment with sqPLM based unsupervised TS models for Sinhala, with our SiTSE dataset as the test set. Our TS models are based on the concept of intermediate task transfer learning (ITTL). In addition to using a single task for the intermediate task as reported in previous research~\citep{phang2020english, phang2018sentence}, we fine-tune the sqPLM sequentially with a set of auxiliary tasks. Given that the type of the intermediate task matters~\citep{phang2020english}, all these auxiliary tasks are sequence-sequence tasks - Sinhala paraphrasing, Translation, and English simplification. 

All our ITTL models report SARI scores comparable to those reported for English datasets~\citep{alva-manchego-etal-2020-data}, as well as high BERTScores. The best SARI result was obtained from the sequential fine-tuning setup with auxiliary task data. Thus we present our work as a case-study on how low-resource languages can exploit existing data and models to achieve acceptable performances on new tasks even without task-specific training data.  

We also conducted two human evaluations on the results and the manually simplified datasets to shed further light on the model performance and data quality. In one evaluation, three evaluators scored the simplifications with respect to fluency, adequacy, and simplicity on a Likert scale of 1-5. In the other evaluation, three separate evaluators recorded the frequency of simplification related errors. Both human analyses confirmed that our models outperform the existing sqPLM-based models for text simplification. Our comprehensive analysis also introduced two new criteria for the human evaluation of text simplification models. We also show that determining the best model with respect to all error analysis criteria is challenging.

\section{Related Work}
\subsection{Text Simplification Datasets}

To train as well as to evaluate text simplification models, a parallel dataset is required where  each parallel pair corresponds to a complex sentence/document and its simpler version. 
Early research on English text simplification used automatically aligned sentences from the complex and simple Wikipedias~\citep{narayan2016unsupervised}.  PWKP~\citep{zhu2010monolingual}, C\&K-2~\citep{kauchak2013improving}, Wikilarge~\citep{zhang2017sentence}, AlignedWL and RevisionWL~\citep{woodsend2011learning} are example corpora automatically compiled from the Wikipedia. However, the suitability of such automatically aligned datasets for text simplification is criticised~\citep{alva-manchego-etal-2020-data}. The TurkCorpus~\citep{xu-etal-2016-optimizing}  was derived by manually simplifying a subset of the PWKP corpus. Simplification was conducted mainly via lexical paraphrasing. ASSET provides 10 simplifications per sentence for a subset of TurkCorpus~\citep{alva-manchego-etal-2020-asset}. HSplit corpus was created by simplifying the test set of the TurkCorpus~\citep{sulem-etal-2018-bleu}. Here, simplification was mainly achieved by sentence splitting.

The SimPA and the Newsela~\citep{xu-etal-2015-problems}~\citep{scarton2018simpa} corpora are manually compiled datasets for English. Newsela~\citep{xu-etal-2015-problems} is an example for a document-level dataset (for English and Spanish). News articles were manually simplified upto five levels, by professionals considering children in different educational levels.  OneStopEnglish~\citep{vajjala2018onestopenglish} is another document-level corpus for English TS.

Table~\ref{table:datasets} lists some of the available datasets for non-English languages. We note that most of the datasets are aligned at the sentence level, and manually aligned datasets have sizes of around 1000. Furthermore, most datasets provide only one reference per complex sentence, which indicates the difficulty of creating large-scale human-curated text simplification datasets.  
\begin{table*}[ht]
\caption{\label{table:datasets} Existing Non-English Text Simplification Datasets}
\resizebox{\textwidth}{!}{
\small

\begin{tabular}{p{0.21\linewidth}p{0.13\linewidth}p{0.1\linewidth}p{0.06\linewidth}p{0.07\linewidth}p{0.13\linewidth}p{0.3\linewidth}}
\hline
\textbf{Paper} & \textbf{Language} & \textbf{Alignment} & \textbf{ Simple Dataset Size} & \textbf{\# of refs} & \textbf{Source} & \textbf{Compilation} \\
\hline
\citet{goto2015japanese} & Japanese & sentence & 13336 & 1 &	news & Instructors simplified full articles. Sentences were manually aligned. \\
\citet{bott2011unsupervised} & Spanish & sentence & 590 & 1	& news & Manualy simplify articles, and sentences were automatically aligned \\
\citet{klaper2013building} & German	& sentence	& 7000 & 1 & websites & Start with complex-simple document pairs and automatically aligned sentences \\
\citet{tonelli2016simpitiki} & Italian	& sentence & 1166 & 1 & wikipedia & By analysing page edits, complex-simple pairs were extracted and later manually curated \\
\citet{brunato2015design} & Italian	& sentence & 1036 &	1 & childrens novel	& Used novels and their manually simplified version, and manually extract parallel sentences \\
\citet{brunato2016paccss} & Italian	& sentence & 63000 & 1 & open & Automatically web-mined \\
~\citet{xu-etal-2015-problems} & Spanish & document & 1221 & 5 & news & Manually simplified by professionals \\
\citet{gala2020alector} & French & document & 79 & 4 & literary, scientific & Manually simplified by professionals \\
\citet{anees2021automatic} & Urdu & sentence & 610 & 2 & literary & Automatically simplified \\
\citet{caseli2009building} & Brazilian Portuguese & sentence & 2116 & 1	& news & Manually aligned \\
\citet{klerke2012dsim} & Danish	& sentence & 48186 & 1 & news & Automatically aligned\\
\citet{brouwers2014syntactic} & French & sentence & 235 & 1 & Wikipedia, literature & First parallel articles were identified, and sentence alignment was done semi-automatically \\
\citet{sauberli2020benchmarking} & German & sentence & 3666 & 2 & news & Start with simple-complex pair of documents. Automatically align sentences in training set and manually align validation and test sets.\\
\citet{specia2010translating} & Portuguese & sentence & 4483 &1  & news &manual simplification\\
\citet{battisti2019corpus} & German & document & 8400 & 1 & news & automatically simplified\\
\citet{miliani2022neural} & Italian & sentence & 736 & - & government & manual simplification\\
\citet{mondal2024dimsim} & Bengali,Marathi & sentence & 500 & 1 &  news & manual simplification\\
\citet{mondal2024dimsim} & Hindi & sentence & 1041 & 1 &  news, Wikipedia & manual simplification\\
\citet{stodden2023deplain} & German & document & 1239 & 1 & news, general & manual and automatic\\
\hline
\end{tabular}}

\end{table*}
{
\subsection{Pre-trained Language Models}
The aim of neural language modeling is to learn a distributed representation over the linguistic units (e.g., words, sentences) of a language~\citep{bengio2000neural}. Neural language models came into prominence with the success of word representation models such as Word2Vec~\citep{mikolov2013distributed}. Later, with the introduction of the Transformer architecture~\citep{vaswani2017attention}, neural language models learned from billions of text tokens came into play, and are referred to as pre-trained Language Models (PLMs) or Foundation Models, among other terms. Based on their training, PLMs can be categorized as encoder-only, decoder-only, and encoder-decoder models. Models such as BERT~\citep{devlin2018bert} and RoBERTA~\citep{liu2019roberta} are examples of encoder-only models. Models such as GPT~\citep{radford2019language}, and even the recently introduced models such as ChatGPT and Llama~\citep{dubey2024llama} (which are commonly referred to as Large Language Models (LLMs)) are examples of decoder-only models. BART~\citep{lewis2019bart} and T5~\citep{raffel2020exploring} are encoder-decoder models, and mBART~\citep{mbart} and mT5~\citep{xue-etal-2021-mt5} (respectively) are their multilingual versions. We call these sqPLMs. Such models have been trained in a seq-seq manner and are suitable for tasks such as TS, which expects to generate a new text string conditioned on an input text string.
}

\subsection{Neural Text Simplification}
The nature of the text simplification task makes the encoder-decoder neural architectures a natural solution, where the simplification task is modeled as a monolingual translation task~\citep{nisioi2017exploring,vu2018sentence}. However,  vanilla encoder-decoder models tend to miss transformations needed for simplification~\citep{zhang2017constrained}.

One way to mitigate this problem is to combine the neural model with a rule-based system~\citep{zhang2017constrained, sulem2018simple,maddela-etal-2021-controllable, zhao2018integrating}. Another solution is to add artificial tokens to the source text to control the simplification process at the decoder~\citep{scarton2018learning,nishihara2019controllable, martin-etal-2020-controllable,mallinson2019controllable}.~\citet{zhang2017sentence,nakamachi2020text} combined reinforcement Learning with the neural model. \citet{kriz2019complexity} added a complexity-weighted loss function to
encourage the model to choose simpler words. Several research exploited the knowledge of related tasks such as entailment, translation and paraphrasing in a multi-task setup~\citep{agrawal2019controlling,guo2018dynamic}. Another line of research incorporated multiple neural models to separately predict the simplifying operation, and execute that edit operation~\citep{dong2019editnts, alva2017learning}.~\citet{bahrainian2024text} employed a secondary neural model to translate complex terms to their simpler versions.

To implement TS systems for languages that have no parallel simplification data, three solutions are available: (1) create simplification data by data augmentation~\citep{martin2021muss,aprosio2019neural}, (2) zero-shot text simplification on a multi-tasking sequence-sequence model that exploits TS data of high-resource languages~\citep{mallinson2020zero} and 3) auto-encoders~\citep{surya2019unsupervised,zhao2020semi}. 

More recent research has used sqPLMs as well as decoder-only PLMs for TS.~\citet{kew2023bless} carried out a comprehensive evaluation across 44 such models and concluded that prompting on decoder-only PLMs (i.e.~LLMs) is the best. However, they only considered English datasets. {

Similar to us,~\citet{martin2021muss, ryan2023revisiting} used sqPLMs for unsupervised TS. However,~\citet{martin2021muss} simply fine-tuned mBART with web-mined paraphrases for the target language. Similar to us,~\citet{ryan2023revisiting} used a TS dataset from another language to fine-tune the sqPLM, but did not consider ITTL.

\subsection{Intermediate Task Transfer Learning (ITTL)}
\label{sec:lit_itft}
ITTL, or Supplementary Training on Intermediate Labeled data Tasks (STILTs) means to fine-tune a PLM with an auxiliary task, before fine-tuning the same with the target task~\citep{phang2018sentence}. This intermediate task can  either be a single task or a multi-task setup~\citep{phang2018sentence}. 

In the context of Encoder-based pre-trained models,~\citet{phang2020english} used ITTL in a cross-lingual zero-shot setup.
~\citet{artetxe2020translation} extensively experimented with the impact of translated data on ITTL. They showed that rather than using English task data as an intermediate task, using its back-translated version performs better. As for sequence-sequence models,~\citet{takeshita2022x} used English monolingual summarization and machine translation as intermediate tasks for multilingual text summarization. For Neural Machine Translation,~\citet{nayak2023leveraging} first fine-tuned a sqPLM with auxiliary domain data before fine-tuning the same with target domain data. In the context of Sinhala,~\citet{dhananjaya2024lexicon} employed ITTL for sentiment analysis.

\section{Sinhala Text Simplification Evaluation Dataset}
\label{sec:sitse}
Addressing the unavailability of a Sinhala text simplification dataset, we present {\bf SiTSE} ({\bf Si}nhala {\bf T}ext {\bf S}implification {\bf E}valuation) dataset, an evaluation dataset for the Sinhala language composed of 1,000 complex sentences, each with three simplified reference sentences (thus 3000 complex-simple pairs).

\paragraph{Data Selection:} Our dataset is from the government domain. In Sinhala, government documents are written in the \textit{written} form of the language, whereas Sinhala speakers use the much simpler \textit{spoken} version of the language for their day-to-day communication. These documents heavily use the government approved standardized term glossaries that contain words that are rarely used in the daily usage of Sinhala speakers. The sentences are also long and the average sentence length of 34.98 words. 

Our data source is the Sinhala side of the English-Sinhala-Tamil parallel corpus prepared by~\citet{fernando2020data}. This dataset has been meticulously cleaned by Sinhala language experts. Three of the authors manually reviewed the corpus and selected 1,000 complex sentences with rare words and large sentence lengths. 
\paragraph{Annotation Process:} Three human participants performed the simplification\footnote{{
Details of human participants are in Appendix~\ref{appendix2}}}. Following  \citet{alva-manchego-etal-2020-asset}, annotators were instructed to perform the following operations to simplify text:

\begin{itemize}
    \item Extract the main idea of the sentence
    \item Split long sentences into shorter ones
    \item Lexical reordering, and replacing complex words with commonly used simple words
\end{itemize}

Annotators were provided 3 rounds of pilots. In each round, 3 of the authors went over the complex-simple sentence pairs, first compiled a list of recommendations, and then provided comprehensive feedback to the annotators. In addition, we periodically monitored and provided feedback to the annotators to ensure they followed the instructions. The annotation was performed in batches, and we monitored each batch to ensure the quality. In Figure~\ref{figure:sitse-sample}, we present an example of simplifications from the SiTSE dataset, having both complex and corresponding simplified sentences. We could see that annotators performed different operations for the simplification.

\begin{figure}[h]
\includegraphics[width=1 \textwidth]{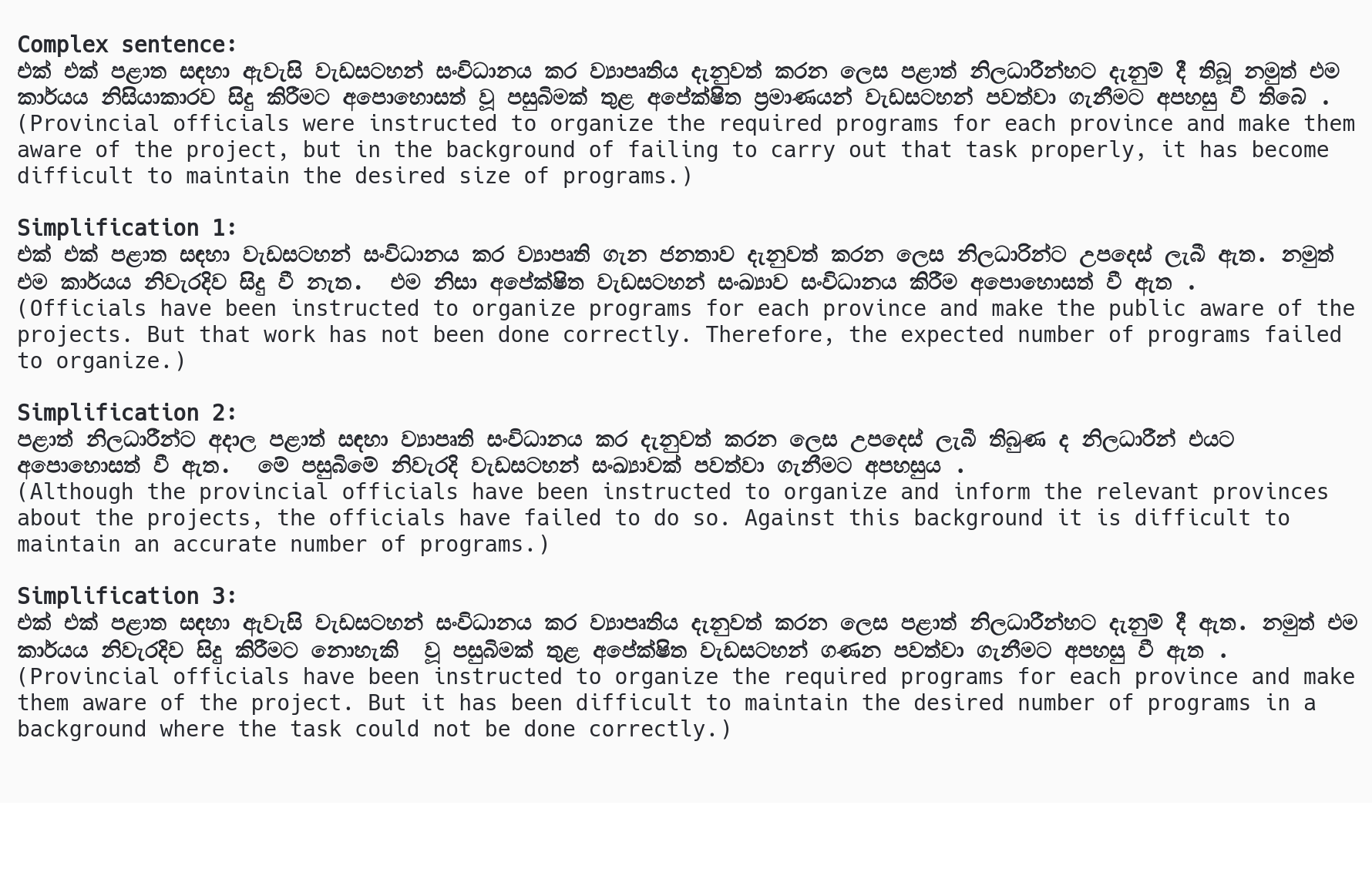}\\
\caption{A sample sentence from SiTSE dataset with English translations}
\label{figure:sitse-sample}
\end{figure}

\paragraph{Dataset Statistics:} 
For each complex sentence in the dataset, there are three simplifications. Thus, for the 1000 complex sentences, there are 3000 simple sentences. On the contrary, we note that most of the datasets reported in Table~\ref{table:datasets} have only one simplification per complex sentence. In Table~\ref{table:sistse-stat}, we present the statistics of the SiTSE dataset, averaged across all the annotators\footnote{We used the implementation in \url{https://github.com/feralvam/easse} for the calculation}. We note that annotators have carried out simplification mainly by splitting long sentences.  Thus, most of the simplified versions have multiple sentences and our dataset can be considered similar to the HSplit dataset~\citep{sulem-etal-2018-bleu}. On average, the annotators have produced shorter sentences, having 11-14 average words per sentence, whereas the original sentence has an average length of 34.98. We provide a human analysis of this dataset in Section~\ref{sec:exp}.

\begin{table}[h]
\centering
\caption{Statistics of the SiTSE dataset. Amount of additions, deletions, splits is calculated by taking ratio of those between original and reference sentence. The compression ratio is calculated by dividing the number of characters in the reference sentence by the number of characters in the original sentence. 
}
\label{eval-dataset-stat} 
\begin{tabular}{llll}
\hline
\textbf{Property} & \textbf{Simple 1} & \textbf{Simple 2} & \textbf{Simple 3} \\
\hline
Avg. Sentence Length & 11.23 & 11.92 & 14.39 \\
Additions & 0.23 & 0.13 & 0.09\\
Deletions & 0.16 & 0.13 & 0.07 \\
Sentence splits & 1.96 & 1.51 & 1.33 \\ 
Compression ratio & 1.03 & 1.00 & 1.02 \\
\hline
\end{tabular}

\label{table:sistse-stat}
\end{table}

\section{Models}
\label{methodology} 
\subsection{Baseline Models}

We used three baseline models based on sqPLMs, as well as a pivot-based model. 

\paragraph{Sequence-sequence Pre-trained Multilingual models (sqPLMs):} Pre-trained mBART and mT5 models were used to generate Sinhala sentences without any explicit fine-tuning for text simplification. Here we input complex sentences into mBART and mT5 and get the output sentences. This is similar to the random baseline.   

\paragraph{Zero-shot text simplification (Si Zero-shot TS):}
The goal is to test the model's performance when there is no data for the target language but other language data is available. First, we fine-tuned sqPLMs with a HRL TS dataset. We then tested it for  Sinhala text simplification. Note that this is the zero-shot method used by~\citet{ryan2023revisiting}. 


\paragraph{Text simplification using paraphrase mining (TS-Mining):} 
This is the unsupervised TS system proposed by~\citet{martin2021muss}. Similar to~\citet{martin2021muss}, first, we mined the paraphrases using LASER embeddings~\citep{artetxe2019massively} for Sinhala sentences in the Si-CC15 monolingual dataset~\citep{dhananjaya2022bertifying}. Then, these paraphrases are used along with controllable sentence simplification~\citep{martin-etal-2020-controllable} to fine-tune a sqPLM for Sinhala TS.

\paragraph{Pivot based text simplification (Pivot):}
Pivoting has been reported as a strong baseline in TS~\citep{martin2021muss}. Here, we assume that the availability of a system for simplification in another language and translation from Sinhala is possible. Specifically, we used Google Translate\footnote{\url{https://translate.google.com/}} to translate Sinhala sentences to English and then used \citet{martin2021muss}'s model for English simplification. Finally, the simplified English sentences are translated back into Sinhala. 


\subsection{ITTL based Models}

In our case, although we now have a human-curated dataset to evaluate Sinhala Text Simplification systems, we do not have parallel data to train a model for the same. This is a classic example of a zero-shot problem. We cast this as a zero-resource text simplification problem by translating a text simplification corpus of an HRL into the Sinhala language, thus creating a synthetic Sinhala simplification dataset. The translated corpus was used to fine-tune the sqPLM. We refer to this as the \textit{Si-simp} model.

Then we extended this model with ITTL. 
For the auxiliary tasks, we consider the following sequence-sequence tasks:

\begin{itemize}
    \item Translation (\textit{Trans}): Fine-tune the sqPLM using a parallel corpus used for Machine Translation with Sinhala on the target side. We hypothesize this task would help the decoder of the sqPLM to generate better Sinhala sentences with good fluency since the decoder sees more Sinhala sentences during fine-tuning.
 
    \item Paraphrasing (\textit{Para}): Fine-tune the sqPLM with paraphrases of the considered language. We used \citet{martin2021muss}'s web mining technique to generate paraphrases automatically. Paraphrasing task helps the model in meaning preservation.
    \item HRL text simplification (\textit{En-simp}): Fine-tune the sqPLM with a text simplification corpus of an HRL similar to our \textit{Si Zero-shot TS} baseline. Fine-tuning with a dataset of the same task teaches task characteristics to the model. As discussed in Section~\ref{sec:lit_itft}, this is a common choice for the intermediate task. 
\end{itemize}

For sequential task fine-tuning, we used combinations of these three auxiliary tasks to determine which task(s)/ordering has the highest positive impact on the Sinhala text simplification. These different modes of ITTL are shown in Figure~\ref{figure:seq-method}. While it is possible to experiment with other auxiliary tasks, we are not aware of any other sequence-sequence datasets for Sinhala.
\begin{figure}[ht]
\includegraphics[width=\textwidth]{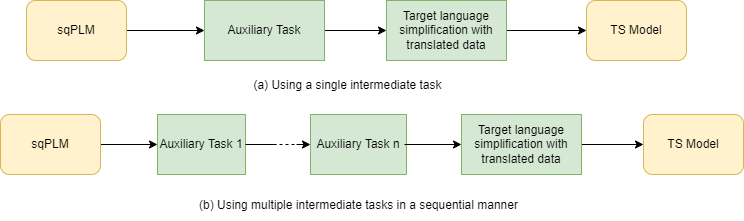}\\
\caption{ITTL with (a) a single intermediate task and (b) a sequence of intermediate tasks.}
\label{figure:seq-method}
\end{figure}

\section{Experiment Setup}

For the translation task, we used the  SiTa corpus with 56,000 parallel Sinhala-Tamil-English sentence pairs extracted from official government documents of Sri Lanka~\citep{fernando2020data}. For the paraphrasing task, we generated 7,000 paraphrase sentence pairs from a portion of 600,000 sentences of the Si-CC15 dataset using \citet{martin2021muss}'s method for paraphrase mining. We limit ourselves to 600,000 sentences from the Si-CC15 dataset due to computational resource limitations. When using this method, we used the Newsela~\citep{xu-etal-2015-problems} dataset for the English text simplification task  as the HRL TS dataset and used Google  Translate to translate the Newsela corpus to Sinhala. 

We used mBART50 and mT5-base models for our experiments. Fairseq~\citep{ott-etal-2019-fairseq} was used for fine-tuning mBART models and HuggingFace Transformer~\citep{wolf-etal-2020-transformers} was used for fine-tuning mT5 models. {
Details on the hyper-parameters and the computation resources are in Appendix~\ref{appendix1}.}

\subsection{Human Evaluation}
\label{sec:humEval}
We employ human evaluators to measure the simplification capability of the humans (see Section~\ref{sec:sitse}) as well as the TS models using \textbf{adequacy} (output sentence preserving the  main idea of the complex sentence), \textbf{fluency} (output sentence free from grammatical errors) and \textbf{simplicity} (output sentence easier to understand than the complex sentence). These are the commonly used criteria for human evaluation on TS~\citep{martin2021muss, maddela-etal-2021-controllable}. We randomly selected 50 sentences from the human-created, as well as the model generated datasets. Three human participants rated the simplified sentence compared to the complex sentence for each of the three criteria and gave a score in a likert scale of 1-5. We calculated the average for each experiment across 50 sentences under each dimension.

In order to further understand the difference in the simplification capabilities of humans and the models, we performed a separate error analysis by using randomly selected 50 sentences and 3 separate evaluators. We used \citet{maddela-etal-2021-controllable}'s error categorization for this purpose. The error categories are: fluency errors, hallucinations, anaphora resolution errors, and bad substitutions. In addition to that, we added two new error categories - near exact copy, and missing major fact. This is because there can be an output sentence that misses a major fact or is same as the complex sentence, while not having any of the other errors. These two error categories are mutually exclusive. {
Further details of these error categories are in Appendix~\ref{appendix3}}.

\section{Results and Discussion}
\label{sec:exp}


\subsection{Quantitative Results}
\label{sec:quantResult}
There are no agreed-upon evaluation metrics for text simplification~\citep{alva-manchego-etal-2020-asset}. Following \citet{alva-manchego-etal-2021-un}'s suggestion, we first computed BERTScore~\citep{bert-score} to measure the quality of the system output and then used SARI to measure the simplicity. {
The higher the SARI and BERTScore, the better the simplification is. More details of these metrics are in Appendix~\ref{appendix4}. }We do not use N-gram based metrics such as  BLEU~\citep{papineni-etal-2002-bleu} because it is not suitable for evaluating text simplification~\citep{sulem-etal-2018-bleu}.{
Recently, several learnable TS metrics have been proposed~\citep{kwak2023context,cripwell2023simplicity,maddela2023lens, zhao2023towards}. However, these metrics are based on a neural model trained using human-annotated data, which is not available for low-resource languages such as Sinhala.}

\begin{table*}[ht]
\caption{Quantitative results. \emph{Si-simp} - Sinhala Simplification,  \emph{En-simp} - English Simplification, \emph{Trans} - Translation, \emph{Para} - Paraphrasing, BERT - BERTScore. `T1\textrightarrow T2' indicates that T1, T2 tasks are trained sequentially. }

\centering

\resizebox{\textwidth}{!}{
\small
\begin{tabular}{llllllllllll} 
\toprule

\multirow{3}{*}{Experiments} & 
\multirow{3}{*}{\begin{tabular}[c]{@{}l@{}}Avg. sent \\ length\end{tabular}} & 
\multicolumn{5}{c}{mBART}  & 
\multicolumn{5}{c}{mT5}  \\ \cline{3-7} \cline{8-12} 
& & 
\multirow{2}{*}{SARI} & 
\multicolumn{3}{c}{Components of SARI} & 
\multirow{2}{*}{BERT} & 
\multirow{2}{*}{SARI} & 
\multicolumn{3}{c}{Components of SARI} & 
\multirow{2}{*}{BERT} \\ \cline{4-6} \cline{9-11}


 &  &  & $F_{ADD}$   & $F_{DEL}$   & $F_{KEPT}$  &   &    & $F_{ADD}$   & $F_{DEL}$   & $F_{KEPT}$  & \\ \cline{1-2}  \cline{3-7} \cline{8-12}
 sqPLMs (Baseline) & 39.12 & 28.80 & 0.19 & 7.40 & 78.81 & 86.20              & 16.35 & 0.08 & 47.00 & 1.99 & 15.21                  \\ 
 Si Zero-Shot-TS (Baseline) & 34.82 & 17.32 & 0.40 & 47.00 & 4.56 & 31.26 & 16.83 & 0.36 & 46.99 & 3.15 & 25.10 \\ 
 TS-Mining (Baseline) & 20.76 & 34.57 & 0.67 & 34.38 & 68.67 & 85.66 & 35.10 & 1.12 & 28.76 & 75.43 & 81.77                  \\ 
 Pivot (Baseline) & 27.74 & 29.12 & 4.87 & 49.47 & 33.02 & 71.05                 & -- & -- & -- & -- & --                  \\ 
 \emph{Si-simp} & 28.12 & \textbf{38.20} & 4.21  & \textbf{38.75} & 71.63 & 81.72                 & 38.39 & 4.16  & 36.47 & 74.54 & 82.90                  \\ 
 \emph{Trans}\textrightarrow \emph{Si-simp} & 29.69 & 37.63 & 4.61  & 35.14 & 73.15 & 83.61                 & 39.17 & 5.25  & 37.91 & 74.37 & 82.94                  \\ 
 \emph{Para}\textrightarrow \emph{Si-simp} & 29.74 & 37.38 & 4.22  & 34.63 & 73.31 & 83.58                 & 38.51 & 4.46  & 37.39 & 73.69 & 82.60                  \\ 
 \emph{En-simp}\textrightarrow \emph{Si-simp} & 29.08 & 37.31 & 4.06  & 35.34 & 72.52 & 82.93                 & 38.37 & 4.79  & 35.64 & \textbf{74.70} & \textbf{83.26}                  \\ 
 \emph{Trans}\textrightarrow \emph{En-simp}\textrightarrow \emph{Si-simp} & 29.50 & 37.81 & \textbf{4.62}  & 35.84 & 72.96 & 83.32 & \textbf{39.95} &\textbf{5.51}  & \textbf{38.28} & 74.25 & 82.85                  \\ 
 \emph{Para}\textrightarrow \emph{En-simp}\textrightarrow \emph{Si-simp} & 30.41 & 37.33 & 4.09  & 33.75 & \textbf{74.14} & \textbf{84.10}                 & 38.63 & 4.71  & 36.74 & 74.46 & 83.05                  \\ 
 \emph{Trans}\textrightarrow \emph{En-simp}\textrightarrow \emph{Para}\textrightarrow \emph{Si-simp}  & 28.91 & 37.65 & 4.53  & 36.25 & 72.18 & 82.94                 & 39.17 & 5.37  & 38.01 & 74.12 & 82.93                  \\
\hline
\end{tabular}
}
\label{table:experiments}

\end{table*}

Results are reported in Table~\ref{table:experiments}. First and foremost, we could observe that BERTScore is comparatively higher (more than 81) for most of the models, except for \emph{Si Zero-Shot-TS} and the \emph{Pivot-based} baselines. As observed by \citet{alva-manchego-etal-2021-un}, a higher BERTScore means good output sentence quality. This means that most models provide good quality sentences - simple or not. Manual inspection showed that the output of \textit{Pivot-based model}\footnote{Adds out-of-context words.} and \emph{Zero-Shot-TS}\footnote{Generates code-mixed data.} baselines are of very bad quality. Therefore  we did not use these two baseline models in further analysis. The rest of the models are analyzed using SARI scores.    

All our ITTL models have comparable results for both mBART and mT5, with mT5 being slightly better. The best result is reported by the  \emph{Trans}\textrightarrow \emph{En-simp}\textrightarrow \emph{Si-simp} ITTL strategy on mT5. We also note that these results are comparable to those reported by~\citet{alva-manchego-etal-2020-data} for English text simplification.  Moreover, all our models are noticeably better than~\citet{martin-etal-2020-controllable}'s model (\textit{TS-mining}).

ITTL that uses translation as the auxiliary task shows the best performance. Similarly, when combining auxiliary tasks, starting with the translation task gives the best result. Thus our observations align with that of~\citet{takeshita2022x}. Combining all three tasks sequentially does not lead to noticeable gains. We also note that SARI scores can sometimes be misleading. For example, the amount of deletions reported in the \textit{TS-mining} baseline is lower than many other models. However, it has the shortest sentence length, which suggests that more content of the original sentence has been removed, which contradicts the corresponding value reported by SARI.  

\begin{table*}[ht]
\centering
\caption{Translation Ablation Result.}
\resizebox{\textwidth}{!}{
\small
\begin{tabular}{lllllllllll} 
\toprule
\multicolumn{1}{c}{\multirow{2}{*}{Experiments}} & \multicolumn{5}{c}{mBART}                             & \multicolumn{5}{c}{mT5}                                \\ \cline{2-11} & \multirow{2}{*}{SARI} & \multicolumn{3}{c}{Components of SARI} & \multirow{2}{*}{BERT} & \multirow{2}{*}{SARI} & \multicolumn{3}{c}{Components of SARI} & \multirow{2}{*}{BERT} \\ \cline{3-5} \cline{8-10}
 &  & $F_{ADD}$   & $F_{DEL}$   & $F_{KEPT}$  &   &    & $F_{ADD}$   & $F_{DEL}$   & $F_{KEPT}$  & \\ \cline{1-6} \cline{7-11}
 \emph{Trans}\textrightarrow \emph{Si-simp} & 37.63 & 4.61  & 35.14 & 73.15 & 83.61                 & 39.17 & 5.25  & 37.91 & 74.37 & 82.94                  \\ 
\emph{Ta-Si translation} & 38.24 & 5.23	& 37.98 & 71.51 & 82.11 & 39.42 & 5.99 &	37.72	& 74.54 &	83.24 \\
\emph{7k dataset} & 37.38 &	4.22 &	34.63 &	73.31	& 83.58 & 38.51	& 4.46 & 37.39 &	73.69 &	82.60\\
\hline
\end{tabular}}

\label{table:ablation-study} 
\end{table*}

\paragraph{Translation Ablation Study} Using translation data provided one of the best performances for sequential training. We explored how the selection of translation data plays a role in performance. For this, we used \emph{Trans}\textrightarrow \emph{Si-simp} in two different settings: using a different source language for translation and different data sizes for translation. For a different source language for translation, we trained with Tamil (Ta)-Sinhala data from the same multi-way corpus~\citep{fernando2020data}. For a different data size, we only used 7k En-Si parallel data for the translation task. This dataset size is comparable to the number of paraphrases we used for~\citet{martin2021muss}'s model. Results in Table \ref{table:ablation-study} show that the source language, and the size of the dataset used in the translation task has a minimal impact on the TS result.

\subsection{Human Analysis}
\label{sec:analysis}
\subsubsection{Human Evaluation}

\begin{table*}[ht]
\centering
\caption{ Human evaluation results of the models and the human-curated dataset. }
\begin{tabular}{lllll}
\hline
\textbf{Model} & \textbf{Adequacy} & \textbf{Fluency} & \textbf{Simplicity} & \textbf{Average} \\ 
\hline
sqPLMs (Baseline) & 4.55 & 3.89 & 1.63 & 3.36 \\
TS-Mining (Baseline) & 3.30 & 3.54 & \textbf{3.21} & 3.35\\
Pivot (Baseline) & 3.51 & 3.55 & 2.27 & 3.11 \\
\emph{Si-simp} & 4.30 & 4.37 & 2.28 & 3.65  \\
\emph{Trans}\textrightarrow \emph{Si-simp} & 4.59 &	4.58 & 1.99 &	3.72 \\
\emph{Para}\textrightarrow \emph{Si-simp}  & 4.38 & 4.34 & 2.05 & 3.59 \\
\emph{En-simp}\textrightarrow \emph{Si-simp} & 4.49 & 4.35 & 1.99 & 3.61 \\
\emph{Trans}\textrightarrow \emph{En-simp}\textrightarrow \emph{Si-simp} & \textit{4.61} &	\textit{4.64} &	1.99 & 3.75\\
\emph{Para}\textrightarrow \emph{En-simp}\textrightarrow \emph{Si-simp} & 4.58 & 4.47 & 2.05 & 3.70 \\
\emph{Trans}\textrightarrow \emph{En-simp}\textrightarrow \emph{Para}\textrightarrow \emph{Si-simp} & 4.48 &	4.54 &	2.27 & \textit{3.76}\\ \hline
Human Simplification & \textbf{4.93} & \textbf{4.7} & \textit{2.89} & \textbf{4.17} \\ \hline
\end{tabular}

\label{table:human-eval}
\end{table*}

\begin{table*}[ht]
\centering
\caption{Results of the error analysis done by humans for the models and human-curated dataset. }

\resizebox{\textwidth}{!}{
\begin{tabular}{lllllll} 
\toprule
\textbf{Model} & \textbf{Hallucinations}& 
\textbf{\begin{tabular}[c]{@{}c@{}}Fluency\\Errors\end{tabular}} &
\textbf{\begin{tabular}[c]{@{}c@{}}Anaphora\\Resolution\end{tabular}}
& \textbf{\begin{tabular}[c]{@{}c@{}}Bad\\Substitution\end{tabular}}&
\textbf{\begin{tabular}[c]{@{}c@{}}Near Exact\\Copy\end{tabular}} &	\textbf{\begin{tabular}[c]{@{}c@{}}Missing Major\\Fact\end{tabular}} \\
\toprule
sqPLMs (Baseline) & 4 &	7.34 & 11.34 & 13.34 & 56.66 & 15.34\\
TS-Mining (Baseline) & 1.33 &	15.34 & 8.66 & 15.34 &	11.34 &	74.66\\
Pivot (Baseline) & 24 & 27.34 & 4.66 & 34.66 & 8 & 38\\
\emph{Si-simp}  & 7.33 & 2 & 4.66 & 2 & 54.66 & 36\\
\emph{Trans}\textrightarrow \emph{Si-simp} & 0 & 3.34 & 1.34 & 0 & 69.34 & 27.34\\
\emph{Para}\textrightarrow \emph{Si-simp}  & 0 & 3.34 & 1.34 & 1.34	& 63.34 & 28\\
\emph{En-simp}\textrightarrow \emph{Si-simp} & 0 & 5.34 & 2.66 & 2 & 59.34 & 24.66\\
\emph{Trans}\textrightarrow \emph{En-simp}\textrightarrow \emph{Si-simp} & 0 & 2 & 0 & 0.66 & 70.66 & 26.66\\
\emph{Para}\textrightarrow \emph{En-simp}\textrightarrow \emph{Si-simp} & 0 & 5.34 &	2 &	2 &	64.66 &	24.66\\
\emph{Trans}\textrightarrow \emph{En-simp}\textrightarrow \emph{Para}\textrightarrow \emph{Si-simp} & 0.67 &	2 &	2.66 &	0 & 62 & 28.66 \\
Human Simplification & 4 & 8 & 2 & 9.34 & 71.34 & 1.34\\ \hline
\end{tabular}
}
\label{table:error-analysis}
\end{table*}

\begin{figure}
\centering
\includegraphics[width=0.9\textwidth]{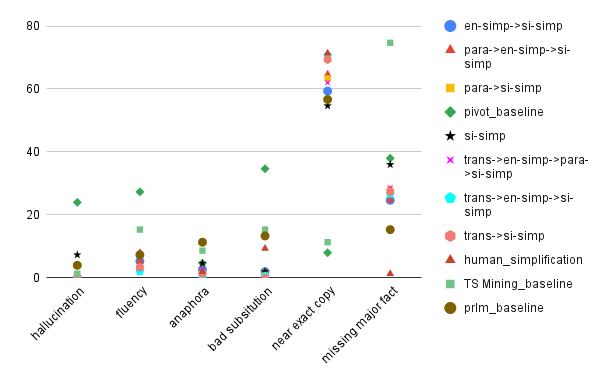}\\
\caption{Visual representation of the number of occurrences of each error type that exist in model outputs and human-curated data.}
\label{figure:error-analysis-chart}
\end{figure}

We start by looking at the `Adequacy', `Fluency', and `Simplicity' of the simplified sentences. In Table~\ref{table:human-eval}, we present the average of the human scores. We observe that the human-curated dataset has the highest values for adequacy, fluency, as well as the average of the three scores. It is also the second best for simplicity, lagging behind TS-mining. However, TS-mining achieves a higher result by simply truncating longer sentences, thus cannot be considered a proper simplification. However, the low value for simplicity relative to that of adequacy and fluency in the human curated dataset is worth investigating. As mentioned in Section~\ref{sec:sitse}, human annotators have mainly focused on sentence splitting when simplifying. However, looking at the complex sentences, most of the words that appear as complex terms are domain-specific terminologies for which a simplified Sinhala term may not exist. This effect is there for model outputs as well - the simplicity score is always lower than the other two scores.

Further, we find that all our sequential training models are better than the baseline models. The best average score is when all the auxiliary tasks are sequenced. However, this gain is insignificant when considering sequential fine-tuning with one and two intermediate tasks. Interestingly, the other top-most results are reported by the ITTL models that have the translation task as the first task. These observations tally with the SARI scores reported above. 

Consistent with the previous research~\citep{kriz2019complexity}, we also observed that the models tend to copy a large portion of the source sentence while generating the output, i.e., the lesser the number of errors, the higher the amount of copying. We leave further investigations to future research.	

\subsubsection{Error Analysis by Humans}

To further deep dive into the model's output, we performed an error analysis based on the six criteria defined in Section~\ref{sec:humEval}. In Table~\ref{table:error-analysis} and Figure~\ref{figure:error-analysis-chart}, we present the average numbers of sentences per error category type. First and foremost, we see the utility of adding two new criteria (`Near exact copy' and `Missing major fact') to the error analysis. Models with low scores for hallucination, fluency, anaphora resolution, and bad substitution tend to have a near exact copy of the complex sentence.

Our error analysis also confirms that our models are better than the baseline models for most cases. In other words, all the baseline models fall into the three worse performing models with respect to different error categories (\textit{pivot} - 5 error categories, \textit{prim} and \textit{TS-Mining}- 4 categories).  

Finally, our comprehensive human evaluation highlights the difficulty of evaluating text simplification systems. The quality of the output of a text simplification system can be analyzed in multiple ways - a model that scores well with respect to one criterion may perform poorly on another criterion. As a consequence, it is difficult to select a single ITTL model as the clear winner of our Sinhala text simplification experiments. Thus we agree with \citet{alva-manchego-etal-2021-un} in need for better metrics for evaluating text simplification systems. However, overall, using ITTL on sqPLMs is a better solution for text simplification than paraphrasing. 

\begin{figure}[ht]
\includegraphics[width=1\textwidth]{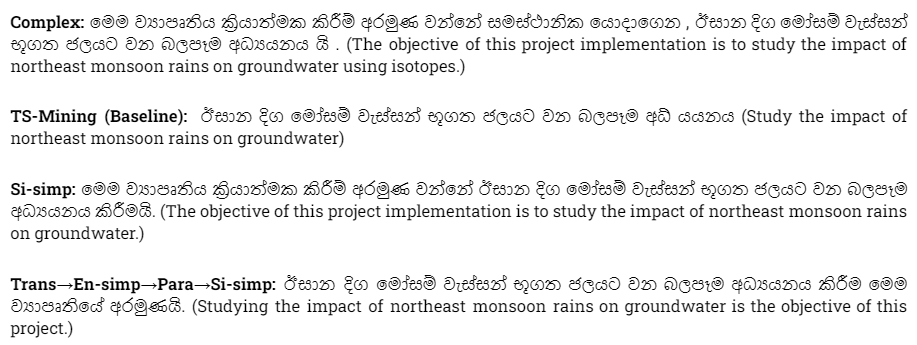}\\
\caption{Sample sentences generated from three models}
\label{figure:sample-sents}
\end{figure}

\subsection{Qualitative Results}
\label{sec:qualitative}
 
 Figure~\ref{figure:sample-sents} shows the sample output generated for the \emph{TS-Mining}, \emph{Si-Simp} and \emph{Trans}\textrightarrow \emph{Si-Simp} models. English translation of each of the sentences is also provided. The sentence generated from the \emph{TS-Mining} baseline is the shortest, and it misses two major points (use of isotopes and the fact that it is a project). \emph{Si-Simp} and \emph{Trans}\textrightarrow \emph{Si-Simp} both have one fact missing. However, the \emph{Trans}\textrightarrow \emph{Si-Simp} model has restructured the sentence in such a way that it appears simpler than the original.


\section{Conclusion}
Text simplification is largely ignored for low-resource languages due to the unavailability of datasets. In response, we created the SiTSE dataset for Sinhala, a language from the Global South known to be low-resource. Our complex dataset is from government documents, and 3 human-curated simplified sentences accompany each complex sentence. We modeled the simplification as a zero-resource problem and tested multiple baseline models using mBART and mT5. Further, we improved the performance using ITTL with the use of different auxiliary tasks. We showed that ITTL outperforms the previous work that used sqPLMs for text simplification. Further, we did a human evaluation and added two new categories for the error evaluation to enrich the existing ones. Our automatic and human evaluations show that we should not rely on only one metric to judge the performance of a model. \\
{
A limitation of this work is the small size of the dataset - since the dataset has only 1000 sentences, we had to use it all for testing, thus leaving nothing for model training. As a solution, we are currently working on creating a much larger TS dataset for Sinhala and some other low-resource languages. This will enable us to experiment with the impact of ITTL by considering TS data from different languages. Moreover, this will enable us to experiment with multi-task learning on sqPLMs for text simplification. With some data for model training, it is also possible to experiment with LLMs such as Llama.}

\section{Acknowledgment}
Data creation was funded by a Senate Research Committee grant of University of Moratuwa. Training of models in this work was made possible by the AWS cloud credits received by Surangika Ranathunga under the \textit{AWS Cloud Credit for Research} program.

\bibliographystyle{ACM-Reference-Format}
\bibliography{sample-base}
\newpage
\appendix
\section{Evaluation Metrics}
\label{appendix4}
{
SARI was introduced by~\citet{xu-etal-2016-optimizing}. SARI compares the output of a TS system against the input sentence as well as the references. It measures the goodness of added/deleted/kept words during text simplification. In SARI, words that are in the output but not in the input sentence are rewarded, if these words occur in the references. The corresponding value $F_{ADD}$ is calculated by taking the n-gram precision and recall of addition operations. Similarly, words that are in both the input and output are rewarded, if these words occur in the references. The corresponding value $F_{KEPT}$ is calculated by taking the precision and recall of the retained n-grams. Deleted words in the output are rewarded if these words are missing in the reference as well. However, $F_{DEL}$ considers only precision, because over-deleting hurts readability much more significantly than not deleting. SARI is the arithmetic average of these values. 
}\\
BERTScore is a reference-based metric that computes the cosine similarity between tokens in a system output and in a manual reference using contextual embeddings \citep{alva-manchego-etal-2021-un}. A higher BERTScore means good output sentence quality.
\section{Error Categories}
\label{appendix3}
\begin{itemize}
    \item \textbf{Fluency Errors} -  The output contains grammatical errors. Repetitions too are included in this category. 
    \item \textbf{Hallucinations} -  The output contains information that was not included in the source sentence. This measures whether the model generates out of context words or phrases.
    \item \textbf{Anaphora Resolution} -  The output contains pronouns that are hard to resolve.
    \item \textbf{Bad substitution} - The output has substituted a phrase that is not simpler than the input.
    \item \textbf{Near exact copy} - The output is the same as the complex sentence, with few common words (e.g.~stop words) changed.
    \item \textbf{Missing major fact }- The output misses a fact that was in the complex sentence, which is important to meaning preservation.
\end{itemize}

\section{Human Participant Details}
\label{appendix2}
Creating the simplified dataset - All annotators were native Sinhala speakers and had a (minimum) undergraduate degree.

First error analysis - We employed three evaluators for this task, not used in the dataset creation task. Two of them have recently completed their undergraduate degree (IT and Architecture) and the third was a final year undergraduate (Engineering), and their primary mode of university education is English.

Second error analysis - We employed three evaluators not used in the above two tasks. One was a Sinhala lecturer, the other was a Sinhala undergraduate, and the third was one of the authors.

All but one human participant (who preferred co-authorship) were compensated adhering to the institution policies. 
\section{Implementation Details}
\label{appendix1}
\paragraph{Hyper-parameters: } 
For fine-tuning  the mBART model we used the adam-optimizer, 3e-05 as the initial learning rate, 0.3 as dropout, 2500 as warm up updates, inverse square root as the learning rate scheduler algorithm, and 32 as the batch size. For the mT5 model we used Adafactor as the optimizer, 3e-05 as the initial learning rate, 0.1 as dropout, 2500 as warm up updates, linear as the learning rate scheduler algorithm, and 4 as the batch size. Due to resource limitations, we used  half precision to fine-tune the mBART model. mT5 experiments were able to run using full precision. 

\paragraph{Computational resources used: } We used AWS \textit{g4dn.2xlarge} instances to carry out the experiments. We estimate that 340 GPU hours on an NVIDIA Tesla T4 were spent on this research for successful fine tuning tasks, paraphrase mining, and erroneous attempts. For paraphrase mining, we used a single GPU instance - an NVIDIA Tesla T4 with 32 vCPUs and 120 GB RAM.

\end{document}